\documentclass[10pt,twocolumn,letterpaper]{article}

\usepackage{icb}
\usepackage{times}
\usepackage{epsfig}
\usepackage{graphicx}
\usepackage{amsmath}
\usepackage{amssymb}
\usepackage{indentfirst}
\usepackage{booktabs}
\usepackage{multirow}
\usepackage{amsmath}
\makeatletter
\newcommand*{\rom}[1]{\expandafter\@slowromancap\romannumeral #1@}
\makeatother

\icbfinalcopy 

\ificbfinal\fi
\begin{document}

\title{NIR-to-VIS Face Recognition via\\ Embedding Relations and Coordinates of the Pairwise Features}

\author{MyeongAh Cho Tae-young Chung Taeoh Kim Sangyoun Lee\\
Image and Video Pattern Recognition Laboratory\\
School of Electrical and Electronic Engineering, Yonsei University, Republic of Korea\\
{\tt\small {maycho0305, tato0220, kto, syleee}@yonsei.ac.kr}
}

\maketitle
\thispagestyle{empty}

\begin{abstract}

NIR-to-VIS face recognition is identifying faces of two different domains by extracting domain-invariant features. However, this is a challenging problem due to the two different domain characteristics, and the lack of NIR face dataset.  In order to reduce domain discrepancy while using the existing face recognition models, we propose a 'Relation Module' which can simply add-on to any face recognition models. The local features extracted from face image contain information of each component of the face. Based on two different domain characteristics, to use the relationships between local features is more domain-invariant than to use it as it is. In addition to these relationships, positional information such as distance from lips to chin or eye to eye, also provides domain-invariant information. In our Relation Module, \textbf{Relation} Layer implicitly captures relationships, and \textbf{Coordinates} Layer models the positional information. Also, our proposed Triplet loss with conditional margin reduces intra-class variation in training, and resulting in additional performance improvements. 

Different from the general face recognition models, our add-on module does not need to pre-train with the large scale dataset. The proposed module fine-tuned only with CASIA NIR-VIS 2.0 database. With the proposed module, we achieve 14.81\% rank-1 accuracy and 15.47\% verification rate of 0.1\% FAR improvements compare to two baseline models.

\end{abstract}

\section{Introduction}
Recently, as Deep Convolutional Neural Network(DCNN) has shown promising performance in computer vision areas and there has been a lot of improvements in the face recognition tasks as well. Specifically, DCNN extracts representative features of the face from an input image and classifies the features into each identity \cite{facenet4, deepface2}. To recognize each identity, there are two methods: Those that use classification layer such as softmax \cite{softmax21} and those that directly learn embedding features that correspond to face similarity such as cosine similarity \cite{facenet4}. Both methods are intended to make large inter-class distance and small intra-class distance which leads to better performance.

Heterogeneous face recognition refers to the face recognition in images acquired from two different domains such as sketch to photo, TIR(thermal infrared) to VIS(visible light), and NIR(near infrared) to VIS \cite{huang2012learning, jin2015coupled, klare2013heterogeneous}. Especially, NIR camera is widely used in video surveillance and security, because in a night or a low-light environment, it is much more useful than the VIS camera \cite{NIR8}. Therefore, there are lots of studies have been done on the NIR-to-VIS among the heterogeneous face recognition.

\begin{figure}[t]
	\centering
	\includegraphics[scale=0.3]{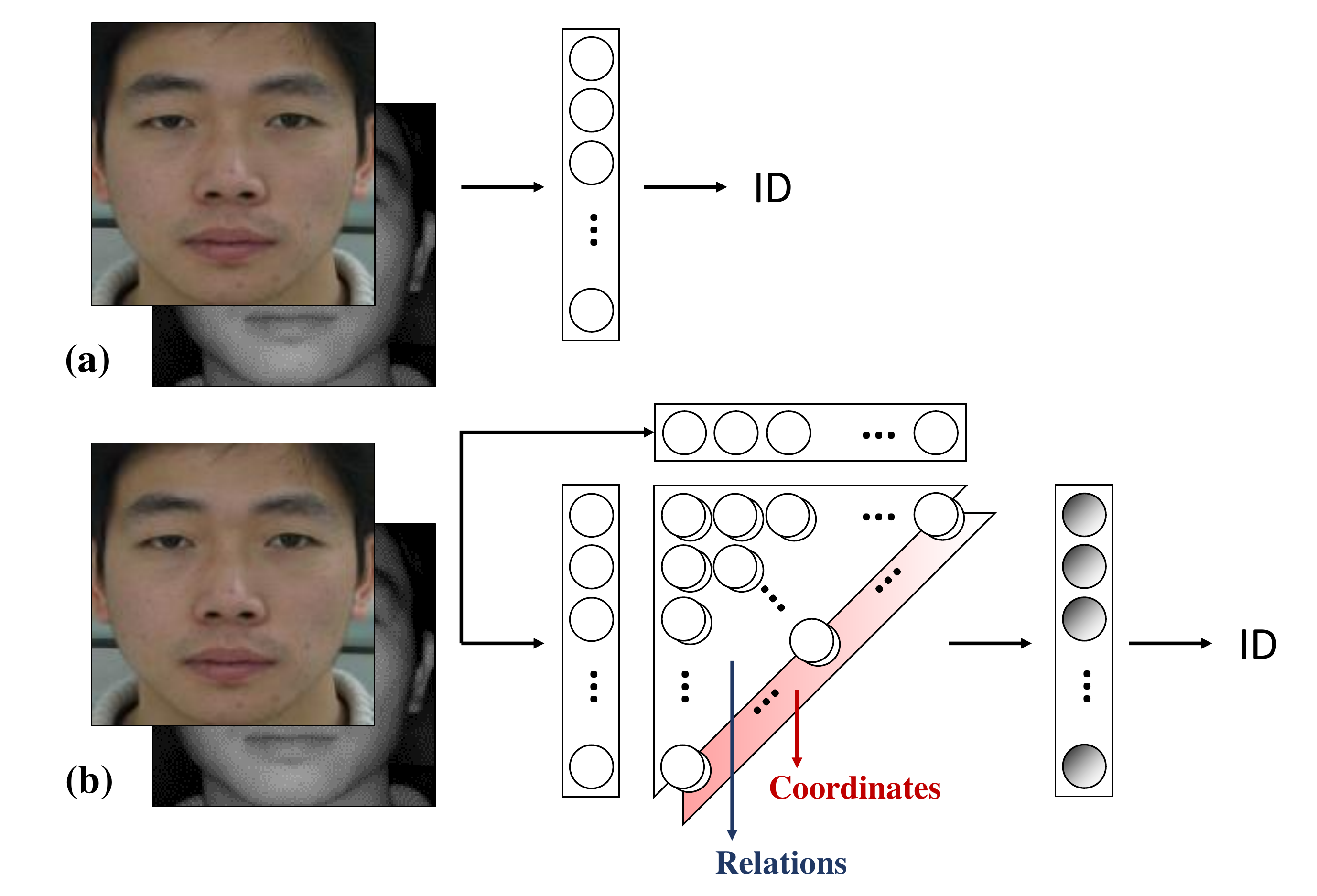}
	\caption{(a) General face recognition network extracts local features with domain discrepancy remaining. (b) Our proposed relation module captures relations and coordinates of the pairwise feature to reduce the domain discrepancy.
	}
	\label{concept}
\end{figure}

The biggest challenging problem of NIR-to-VIS face recognition is extracting domain-invariant features. Therefore, it is important to enlarge each inter-class variation and reduce the discrepancy between the NIR and VIS features in each intra-class. In He \textit{et al.} \cite{HFR26}, network extracts domain-invariant features, Liu \textit{et al.} \cite{HFR24} uses triplet loss with hard sampling and Song \textit{et al.} \cite{GAN11} uses CycleGAN \cite{GAN12} to convert NIR to VIS face images.




In face recognition, it is very important to extract a general feature that can distinguish each person. Face recognition networks use the large scale datasets such as MS-Celeb-1M \cite{ms19} or Labeled Faces in the Wild \cite{huang2008labeled} to generalize features. However, the NIR-to-VIS dataset is a relatively small dataset for training, so the network which trained only NIR-to-VIS dataset cannot provide satisfying performance. Therefore, most of the NIR-to-VIS studies were done by fine-tuning the pre-trained models which is difficult to design a new architecture or transfer learning to a face recognition model with a good performance.

In order to solve these problems, as mentioned above, this paper proposes an add-on ‘Relation Module’ that exploits off-the-shelf models trained on the visual data to extract domain-invariant features without pre-training procedure. Since texture information is dominant in the domain difference, our add-on module only extracts the relationships between texture information. Inspired by the Relation Network \cite{rel15}, our module captures relationships of face components. It modeled the relationship between each object in the image and applied to the Visual Q\&A problem. Similarly, our proposed Relation Module reduces the domain discrepancy between the NIR and VIS by capturing the relationships between each component of the face. After passing through convolutional network, each cell of the feature map represents local patch of the input face image such as lips, eyes and chin. Our module looks all possible combinations of them and does not need to indicate actual relation of patches explicitly. Since these \textbf{relations} represent characteristic of identities and domain-invariant, it can be suitable for the NIR-to-VIS recognition task. Positional information of each component is also important information. Such as distance from lips to chin or eye to eye can be characteristics of identity. Since it is also domain-invariant information, we add \textbf{coordinates} layer to instruct the positions on the Relation Module.

To reduce an additional domain discrepancy, we propose triplet loss with conditional margin. A conditional margin is proposed to provide adaptive margin to the intra-class. In addition, following \cite{HFR24}, anchor and positive, negative examples are sampled from the different domains.

In this paper, our main contributions are,
\begin{itemize}
	\setlength{\itemsep}{0.5em}
	\item To reduce discrepancy between the two different domain features, Relation Module captures relationships and positions between the pairwise patches that are the components of the face. 
	\item This paper proposes triplet loss with conditional margin which considers intra-class distribution. Also for hard sampling, all anchors and targets(positives, negatives) are sampled from the different domain.
	\item Add-on module shows 14.81\% performance improvement over baseline without pre-training and 4.19\% performance improvement over fully pre-trained baseline models. Our maximum rank-1 accuracy achieves 98.92\% which is comparative with the state-of-the-art methods.
\end{itemize}
\begin{figure*}[t]
	\centering
	\includegraphics[scale=0.6]{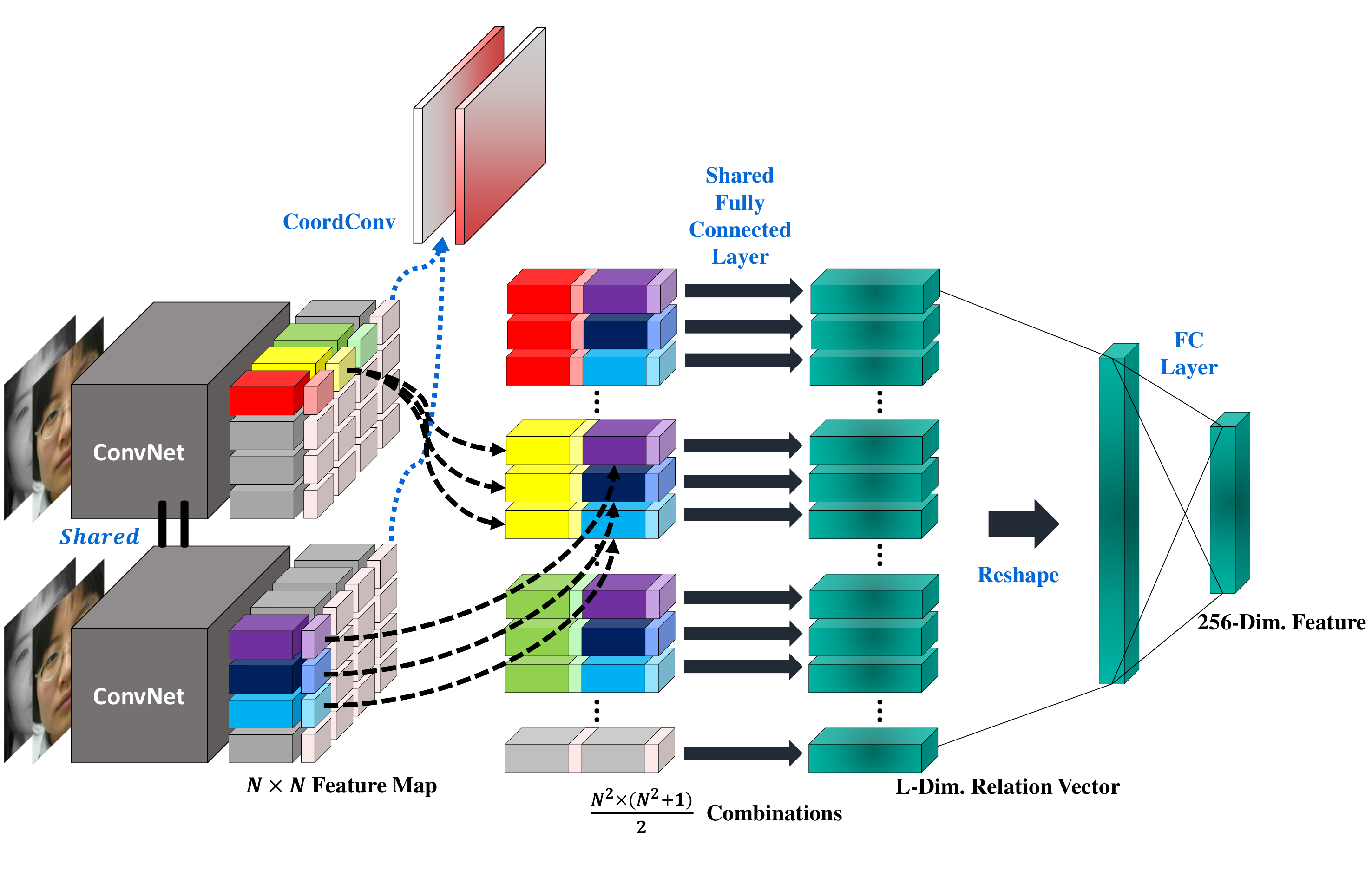}
	\caption{Framework of the proposed Relation Module. It concatenates every pairwise combinations of the features from the ConvNet and the position map. The concatenated vectors are embedded by the shared fully connected layer into the final 256-dimension feature vector.
	}
	\label{network}
\end{figure*}
\section{Related Works}
Conventional approaches for the NIR-to-VIS face recognition is modality-invariant feature learning, which learn features that can create a robust feature space with two different modalities. With the development of deep learning method, Yi \textit{et al.} \cite{yi2015shared} resorts to RBM combined with removed PCA feature. Liu \textit{et al.} \cite{HFR24} improves performance by applying CNN with triplet loss concepts to the NIR-to-VIS face recognition. Wu \textit{et al.} \cite{wu2017coupled} utilizes the low rank and block-diagonal constraints on fully connected layer to alleviate overfitting, and proposes  cross-modal ranking to reduce domain discrepancy. He \textit{et al.} \cite{he2018wasserstein} uses the Wasserstein distance \cite{arjovsky2017wasserstein} to reduce the domain gap, in order to obtain domain-invariant features for the NIR-to-VIS face recognition.

Another approach is utilizing data synthesis which transforms face images from one modality into another, via image synthesis. Data synthesis is first proposed to synthesize and recognize a sketch image from a face photo in Wang \textit{et al.} \cite{wang2009face}. After development of deep learning and GAN \cite{goodfellow2014generative}, Zhao \textit{et al.} \cite{zhao2017dual} performed data synthesis using GAN and Song \textit{et al.} \cite{GAN11} utilized CycleGAN \cite{GAN12} to realize a cross-spectral face hallucination, facilitating heterogeneous face recognition via generation.

Our Relation Module is inspired by Santoro \textit{et al.} \cite{rel15}, which proposed a relation network that finds the relationships between objects in Visual Q\&A problem. Kang \textit{et al.} \cite{kang2018pairwise} also applied the same concept of the relation network for face recognition. However, we applied it to the NIR-to-VIS task because the relation network operation has domain-invariant characteristics.

Liu \textit{et al.} \cite{coord17} has found that directly encoding the positional information of the features under certain circumstances, can be very useful for improving the performance of the network. Vaswani \textit{et al.} \cite{vaswani2017attention} also improved performance by designing encoders and decoders with a positional encoding, so that the network can utilize sequential order information. Therefore, this direct positional information inspired the design of the coordinates layer, which made it  useful for the Relation Module to find the relation between each feature.

Various studies on face recognition focused on loss design to increase discriminative power of the feature, which can be divided into softmax based and triplet based loss methods. In recent years, studies using angular margins based on cosine similarity, such as \cite{deng2018arcface, liu2017sphereface, wang2018cosface}, allowed the face recognition performance to be improved. These studies can also be applied to NIR-to-VIS face recognition. Though Liu \textit{et al.} \cite{HFR24} applied triplet loss to the NIR-to-VIS face recognition, the application of other improved losses with the concept of angular margins to the NIR-to-VIS has not yet been studied sufficiently.



\section{Proposed Approach}

In this section we present an overview of our network and method of the proposed Relation Module that consist of the relation layer and coordinates layer. Then, triplet loss with conditional margin and hard sampling is introduced, which reduces gap between NIR and VIS face images.

\subsection{Overview}
Our network is designed to learn extracting similar embedding features from different domains of face images. The whole framework is illustrated in Figure \ref{network}. Input of the network is NIR or VIS face image and after ConvNet, \textit{N$\times$N} feature map is extracted. For feature extractor baseline, we use LightCNN \cite{light16}. This feature map is input of the Relation Module. In Relation Module, we consider \textit{N$\times$N} number of the feature vectors and all pairwise combinations of them with positional information. These pair sets of combinations pass through shard fully connected layer and are embedded to the relation vector with \textit{L}-dimension. After fully connected layer, we finally extract 256-dimension embedding feature vector which represents each identity.

During training time, we use softmax classifier-based method and triplet loss with conditional margin which is embedding-based method. For triplet loss, we sampled anchor in one domain and negative, positive examples from the other.

\subsection{Relation Layer}
From the fact of that CNN has local connectivity characteristic, each cell of the feature maps after CNN represents the local parts of the input, and each channel-wise vector holds representative information within local parts. 

In Figure \ref{network}, the output of ConvNet is \textit{N$\times$N} feature map(we use \textit{N}=8). These \textit{N$\times$N} number of feature vectors represent the local patches of the face such as lips, eyes and nose which are important characteristics of the face. In the relation layer, we consider all pairwise combinations of the feature vector. By pairwise combining, the relations between two patches of the face can be obtained. Since these relations are regardless of the order, there are \textit{2N$\times$(2N+1)/2} orderless combinations. These combinations are embedded into the \textit{L}-dimension relation vector by shared fully connected layer. This process extracts representative relations of patches such as relation of shapes, size, etc. Relation layer does not need to define explicit or actual relation but simply looks all combinations of patches and discovers general relation implicitly. Since these relations reduce domain dependency, each identity is represented as similar relation vector regardless of the domain.
 
\subsection{Coordinates Layer}
Position of each part of the face is an important information while classifying faces. Relative distance of face parts such as distance from lips to chin or eye to eye can be the representative features of the identities. Since this information is not dependent on the domain, it can be effectively used for NIR-to-VIS face recognition task. Therefore, we add coordinates layer to each feature vector that can give positional information of each patch. Similar to \cite{coord17}, we simply add two additional channels which indicate two spatial dimensions. The first of the first channel is filled with 0’s, second row is filled with 1’s, etc. Second channel is also filled similarly to the first, but columns are constant value and are scaled to [-1,1]. As depicted in Figure \ref{network}, these coordinates(CoordConv) are concatenated with each vector and used in capturing relations.

\begin{table*}[!t]
	\begin{center}
		\begin{tabular}{c|c|c|c|c|c}
			\multicolumn{1}{c}{} &       & Rank-1 Acc.(\%) & VR@FAR=1\%(\%) & VR@FAR=0.1\%(\%) & VR@FAR=0.01\%(\%) \\
			\midrule
			\midrule
			\multicolumn{2}{c|}{Pre-trined model \rom{1}} & 93.21 & 98.01 & 93.41 & 90.15 \\
			\midrule
			\multicolumn{2}{c|}{Baseline \rom{1}} & 82.59 & 93.9  & 80.87 & 74.62 \\
			\midrule
			\multicolumn{2}{c|}{+ Relation Layer} & 94.73 & 98.02 & 93.65 & 91.28 \\
			\midrule
			\multicolumn{2}{c|}{+ Coordinates Layer} & 95.21 & 98.09 & 94.46 & 91.52 \\
			\midrule
			\multicolumn{2}{c|}{+ Conditional Triplet} & \textbf{97.4} & \textbf{99.2} & \textbf{96.34} & \textbf{94.31} \\
			\specialrule{.12em}{.1em}{.1em}
			\multicolumn{2}{c|}{Pre-trined model \rom{2}} & 97.65 & 99.34 & 97.79 & 96.84 \\
			\midrule
			\multicolumn{2}{c|}{Baseline \rom{2}} & 95.21 & 97.85 & 93.83 & 91.36 \\
			\midrule
			\multicolumn{2}{c|}{+ Relation Layer} & 98.12 & 99.37 & 97.68 & 96.86 \\
			\midrule
			\multicolumn{2}{c|}{+ Coordinates Layer} & 98.59 & 99.23 & 97.59 & 96.69 \\
			\midrule
			\multicolumn{2}{c|}{+ Conditional Triplet} & \textbf{98.92} & \textbf{99.44} & \textbf{98.72} & \textbf{98.30} \\
		\end{tabular}%
	\end{center}
	\caption{Results of the proposed method from the baseline on the 10-fold CASIA NIR-VIS 2.0 database.}
	\label{tab:addlabel}%
\end{table*}%
 	
\subsection{Loss Function}
\subsubsection{Softmax Loss}
While training the network, we use softmax classification loss and triplet loss. For softmax loss, we normalize the embedding feature $x_{i}$ by \textit{L2} normalization which is followed by \cite{L2softmax22, softmax21}. Also, normalized feature is re-scaled to scale $s$ followed by \cite{L2softmax22}. In Equation \ref{e2}, we denote batch size $N$, the number of class $M$, weights of the last softmax layer $w$ and the embedding vector $\hat x$.

\begin{gather}
	{\hat x}_{i} = \frac{x_{i}}{\left \| x_{i} \right \|}\times s \nonumber\\
	L_{Softmax} = \frac{1}{N}\sum_{i}^{N}\frac{e^{w_{i}^{T}{\hat x}_{i}+b_{i}}}{\sum_{j}^{M}e^{w_{j}^{T}{\hat x}_{j}+b_{j}}}
	\label{e2}
\end{gather}
\subsubsection{Triplet Loss with Conditional Margin}
Since there are large intra-class discrepancies between two domain differences, triplet loss is introduced. Equation \ref{e3} is original triplet loss function \cite{facenet4}, where $x^{a}$(anchor) is the embedding feature vector of the randomly selected input image and $x^{p}$(positive) is the embedding feature vector of the same class with anchor while $x^{n}$(negative) is the different class with anchor. Loss function is designed to minimize the Euclidean distance between same identity, and to maximize the distance between different identities. 

\begin{figure}[t]
	\centering
	\includegraphics[scale=0.6]{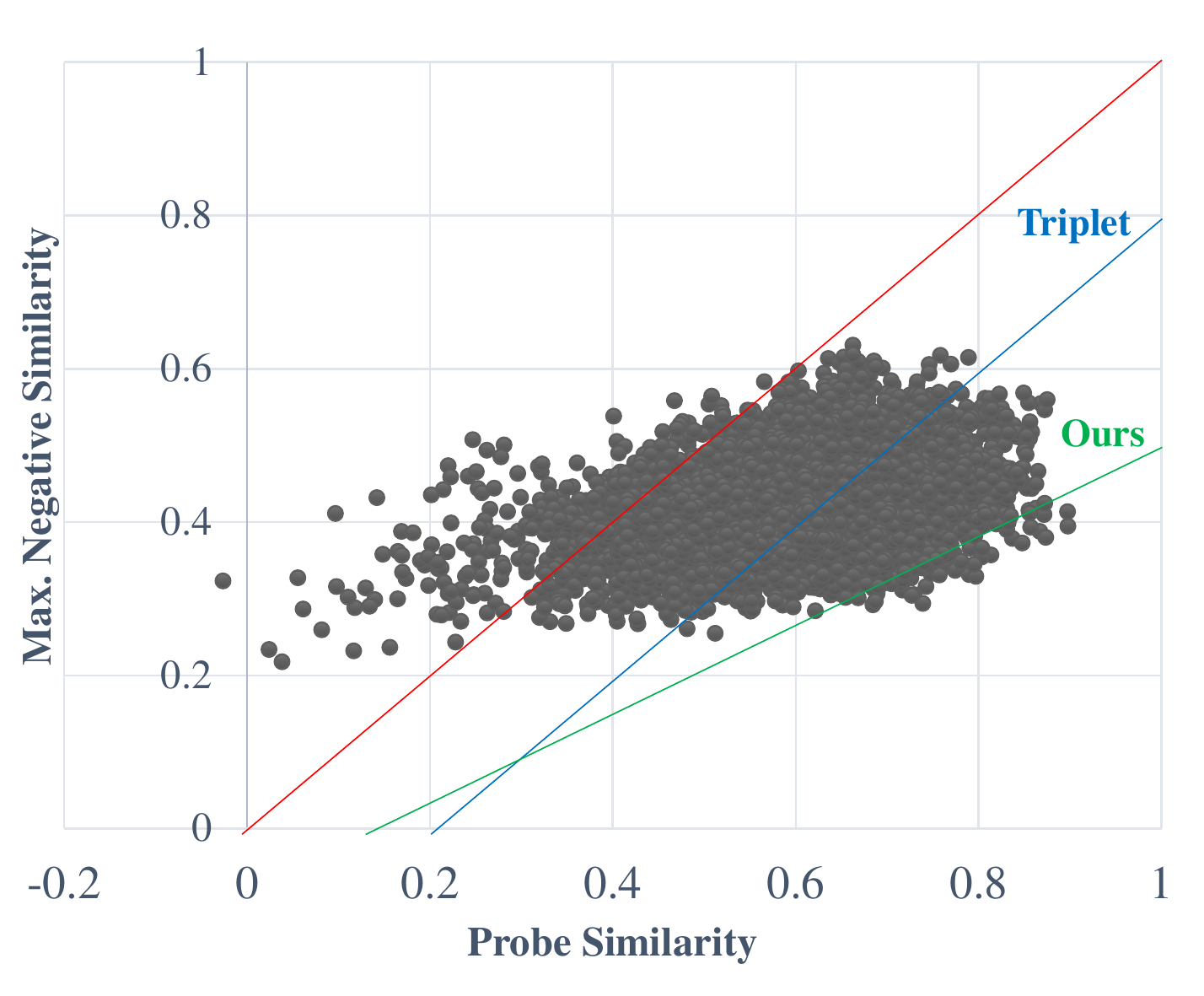}
	\caption{Distribution of the probe similarity versus maximum negative similarity of the CASIA NIR-VIS 2.0 training set. With or without triplet loss, the margin constraint pushes every samples identically while our conditional margin gives adaptive margin which follows the correlation of the distribution.
	}
	\label{triplet}
\end{figure}

\begin{gather}
	\left\{ x_{i}^{a}, x_{i}^{n}, x_{i}^{p}\right\}\in \textit{T} \nonumber\\
	L_{Triplet} = \sum_{i}^{N} [ \left \|x_{i}^{a} -x_{i}^{p} \right \|_{2}^{2} \nonumber\\
	- \left \| x_{i}^{a} - x_{i}^{n} \right \|_{2}^{2}+m ]_{+}
	\label{e3}
\end{gather}

In Equation \ref{e3}, the distance difference should be bigger than margin $m$. We plot the closest negative embeddings for each class in Figure \ref{triplet}. X-axis is cosine similarity between anchor-positive($Sp$) and Y-axis is the maximum cosine similarity of anchor-negative($Sn$) in the training set. However, Equation \ref{e3} is inappropriate for our task. As we can see in this plot, correlation between $Sp$ and $Sn$ is not 1 which means loss criterion should be different according to each $Sp$. Considering $Sp$ and $Sn$ distribution, we propose conditional margin. 
\begin{gather}
S_{p} = CS(x_{i}^{a}, x_{i}^{p})\nonumber\\
S_{n} = CS(x_{i}^{a}, x_{i}^{n})\nonumber\\
\frac{S_{n}+1}{S_{p}+1} < m\nonumber\\
L_{Conditional} = \sum_{i}^{N}\left [ \frac{S_{n}+1}{S_{p}+1}-m \right ]_{+}
\label{e4}
\end{gather}

In Equation \ref{e4}, CS represents Cosine Similarity and we apply conditional margin that considers margin adaptively. In Figure \ref{triplet}, the triplet loss with conditional margin line is considered not only intercept  value $(1-m)$ but also slope $m$. Equation \ref{e4} is our triplet loss with conditional margin ($m$=0.7) and total loss is defined in Equation \ref{e5}.

\begin{equation}
L = L_{Softmax} + \lambda L_{Conditional}
\label{e5}
\end{equation}

To reduce gap between domains, we sampled positive and negative examples at different domains with an anchor \cite{HFR24}. This sampling forces NIR and VIS embeddings to be close having same identity and makes compact intra-class regardless of the domain. 


\section{Experiments and Results}
\subsection{Database}
For experiment, we use CASIA NIR-VIS 2.0 Face Database \cite{casia18}. This database consists of 725 identities and 10 fold experiments. There are 1-22 VIS images and 5-50 NIR images per identities. It is the largest and challenging database of heterogeneous task. We cropped each image by 144$\times$144 size and during training we randomly crop to  128$\times$128 size. In training set, there are about 8,600 number of NIR or VIS images from 360 identities. In test set, gallery set consists only one VIS image and probe set consists about 6,000 NIR images from 358 identities. 

\subsection{Implementation}
Our baseline is LightCNN(removed softmax layer) and it has 9(or 29) number of convolutional layers. This baseline is pre-trained on MS-Celeb-1M dataset \cite{ms19}. Relation Module gets 8$\times$8 size feature map for the input and embedding to 64-dimension relation vectors. Our Relation Module is only fine-tuned with the CASIA NIR-VIS 2.0 database. To prevent classifier overfitting on the training set, we apply dropout at the last softmax layer. Learning rate starts from ${10}^{-3}$ and gradually drops to ${10}^{-5}$. The batch size is set to be 128 and balancing parameter $\lambda$ is 10.

\subsection{Results}
\subsubsection{Relation Module Results}
We followed CASIA NIR-VIS 2.0 Face Dataset View 2 evaluation protocol which consists of 10 sub experiments. All experiments identities in training set and test set are non-overlapping. Table \ref{tab:addlabel} shows results of Rank-1 identification rate and verification rate of 1, 0.1 and 0.01\% FAR. In Table \ref{tab:addlabel}, pre-trained model \rom{1} is LightCNN-9 which indicates whole network is pre-trained on MS-Celeb-1M and fine-tuned on CASIA NIR-VIS 2.0 database. Baseline \rom{1} is LightCNN-9 which only the feature extractor is pre-trained and fully connected layers are fine-tuned. Pre-trained model \rom{1} rank-1 accuracy is 93.21\% and baseline \rom{1} is 82.59\%. We add relation layer and coordinates layer to the baseline \rom{1} where FC layers are removed. The results of the relation layer is 94.73\% and addition of the coordinates layer reaches 95.21\%. Futhermore, we add triplet loss with conditional margin which performed 97.4\%, showing 14.81\% accuracy improvement from the baseline \rom{1}. Since Relation Module does not need pre-training, any other face recognition feature extractor can be added with simple fine-tune procedure. In Table \ref{tab:addlabel}, for pre-trained model \rom{2} and baseline \rom{2}, we use LightCNN-29 which has 29 convolutional layers. Pre-trained model \rom{2} and baseline \rom{2} performed 97.65\% and 95.21\%. After adding relation module and triplet loss with conditional margin, 98.92\% accuracy and 98.72\% verification rate of 0.1\% FAR.

\begin{table}[t]
	\begin{center}
		\begin{tabular}{c|c|c}
			& Rank-1 Acc.(\%) & VR@FAR=0.1\%(\%) \\
			\midrule
			\midrule
			HFR-CNN\cite{HFR25} & 85.9  & 78 \\
			\midrule
			COTS  & \multirow{2}[2]{*}{89.59} & \multirow{2}[2]{*}{-} \\
			+Low-rank\cite{lowrank23} &       &  \\
			\midrule
			TRIVET\cite{HFR24} & 95.7  & 91 \\
			\midrule
			IDR\cite{HFR26}   & 97.33 & 95.73 \\
			\midrule
			ADFL\cite{GAN11}  & 98.15 & 97.21 \\
			\midrule
			CDL\cite{wu2017coupled}   & 98.62 & 98.32 \\
			\midrule
			W-CNN\cite{he2018wasserstein} & 98.7  & 98.4 \\
			\midrule
			Ours  & \textbf{98.92} & \textbf{98.72} \\
		\end{tabular}%
	\end{center}%
	\caption{Comparing with other deep learning based methods on the 10-fold CASIA NIR-VIS 2.0 database.}
	\label{compare}
\end{table}%

\begin{table}[h]
	\begin{center}
		\begin{tabular}{c|c|c}
			& margin(slope) & Rank-1 Acc(\%) \\
			\midrule
			\midrule
			Softmax & -     & 95.21 \\
			\midrule
			Softmax+Triplet & 0.2   & 94.74 \\
			\midrule
			\multicolumn{1}{c|}{\multirow{3}[5]{*}{\shortstack{Softmax\\+Triplet with \\conditional margin}}} & 0.4 (0.6)   & 94.94 \\
			\cmidrule{2-3}          & \textbf{0.3 (0.7)}   & \textbf{97.4} \\
			\cmidrule{2-3}          & 0.2 (0.8)   & 95.62 \\
		\end{tabular}%
	\end{center}
	\caption{Results on the 10-fold CASIA NIR-VIS 2.0 database with the different loss functions and the margin values.}
	\label{tab:addlabel2}%
\end{table}%

As described in Table \ref{compare} shows other HFR models on CASIA NIR-VIS 2.0 database based on deep learning. Our comparing models are HFR-CNN(2016) \cite{HFR25}, COTS+Low-rank(2017) \cite{lowrank23}, TRIVET(2016) \cite{HFR24}, IDR(2017) \cite{HFR26}, ADFL(2018) \cite{GAN11}, CDL(2017) \cite{wu2017coupled} and W-CNN(2018) \cite{he2018wasserstein}. Experiment results are shown in Table \ref{compare}. Our Relation Module with conditional triplet loss performed 0.52\% improvements than W-CNN, which is comparative results with the state-of-the-art models. 

\begin{figure}[t]
	\centering
	\includegraphics[scale=0.43]{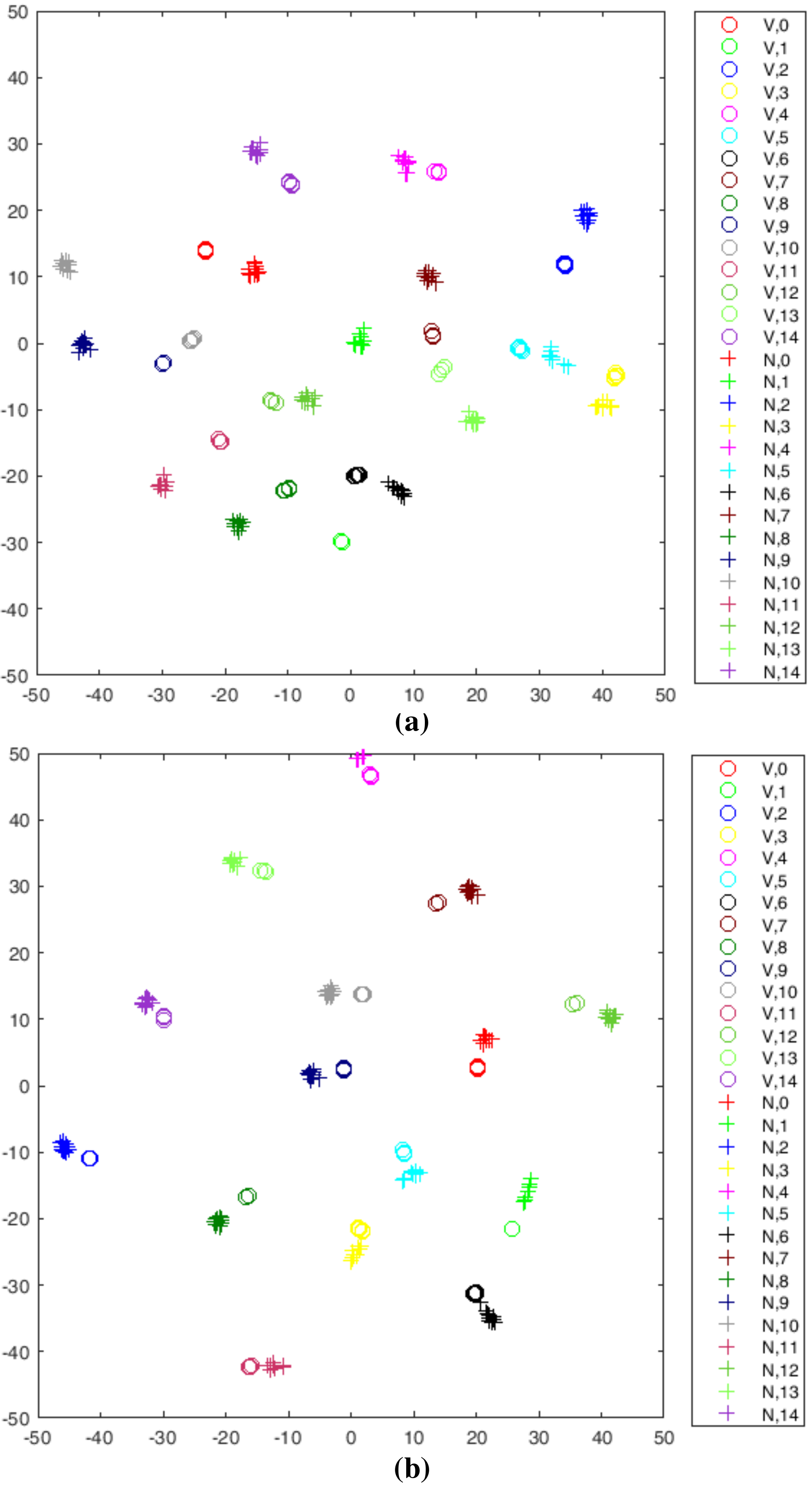}
	\caption{t-SNE Visualization of 256-dimension embedding features (a) in the baseline and (b) in the proposed module.}
	\label{visualization}
\end{figure}

\subsubsection{Triplet Loss with Conditional Margin Results}
We applied different loss in Relation Module network(baseline is LightCNN-9). When we only use softmax loss, rank-1 accuracy is 95.21\% in Table \ref{tab:addlabel2}. In Table \ref{tab:addlabel2}, we experiment $m$ = 0.6, 0.7 and 0.8 representing the values of slope, 0.6, 0.7 and 0.8 and intercept, 0.4, 0.3 and 0.2. Comparing original triplet margin with our suggested conditional margin, proposed loss brings performance gain for 2.38\% in rank-1 accuracy. By modifying $m$, it performs the best when $m$ = 0.7, of 97.4\%. This shows that network needs to train with enough conditional margin and is database dependent parameter. This result shows that minimizing intra-class with conditional margin while separating inter-class with softmax loss is more effective in training.

\subsubsection{Visualizaion of Embedding Features}
We visualize output 256-dimension embedding feature vectors of the trained network which input is NIR and VIS face images. By following t-SNE \cite{maaten2008visualizing}, shown in Figure \ref{visualization}, \textit{V} and \textit{N} indicates VIS and NIR embeddings and the number denotes each identity. (a) is a result of the baseline and (b) is a result of our proposed module which adds Relation Module and triplet loss with conditional margin. In Figure \ref{visualization} each color indicates the identities and most of the identities in (b) distinguish-ably separated. Also comparing distance between NIR and VIS within identity, (b) is much closer than (a) showing compact intra-class. For example, in (a), \textit{V10} and \textit{N10} (or \textit{V13} and \textit{N13}) are distanced each other and close to other identity which leads wrong identification. While in (b), most of embedding features of each class are compact and all identities are separated enough which leads good performance. 

\section{Conclusion}



Relation Module, an add-on module, which simultaneously captures relations and coordinates of the pairwise features from the off-the-shelf models was proposed in this paper. The relation layer effectively captures the pairwise relationship of each component of the face, and the
coordinates layer models the positional information from the features. Furthermore, the proposed triplet loss with conditional margin increases the performance by modeling data dependent adaptive margin between the anchor-positive and the anchor-negative.

Experimental results show that each component of our Relation Module increases the accuracy by only training on the target dataset from the baseline models, showing the comparative performance with the state-of-the-art algorithms. Our visualization of the embedding feature shows
that the Relation Module effectively not only reduce the domain discrepancy between the NIR and VIS but also enlarge the relative inter-class distances.

One of the main difficulties of the heterogeneous face recognition is the lack of the labeled dataset from different domains. The proposed method effectively can solve this problem by combining existing visual face recognition model with small size NIR-VIS face dataset. Our future works will be extended with the same framework to other domains such as sketch and thermal.

\section*{Acknowledgement}
This research was supported by Multi-Ministry Collaborative R\&D Program(R\&D program for complex cognitive technology) through the National Research Foundation of Korea(NRF) funded by MSIT, MOTIE, KNPA(NRF2018M3E3A1057289) 

This work was supported by Institute of Information \&
communications Technology Planning \& Evaluation (IITP) grant funded by the Korea government(MSIT) (2016-0-00197, Development of the high-precision natural 3D viewgeneration technology using smart-car multi sensors and deep learning)

{\small
\bibliographystyle{ieee}
\bibliography{egbib}
}

\end{document}